# Classifying Cue Phrases in Text and Speech Using Machine Learning

Diane J. Litman
AT&T Bell Laboratories
600 Mountain Avenue, Room 2B-412
Murray Hill, New Jersey 07974
diane@research.att.com

## Abstract

Cue phrases may be used in a *discourse* sense to explicitly signal discourse structure, but also in a *sentential* sense to convey semantic rather than structural information. This paper explores the use of machine learning for classifying cue phrases as discourse or sentential. Two machine learning programs (CGRENDEL and C4.5) are used to induce classification rules from sets of pre-classified cue phrases and their features. Machine learning is shown to be an effective technique for not only *automating* the generation of classification rules, but also for *improving* upon previous results.

## Introduction

*Cue phrases* are words and phrases that may *sometimes* be used to explicitly signal discourse structure. For example, when used in a *discourse* sense, the cue phrase "incidentally" conveys the structural information that a topic digression is beginning. When used in a *sentential* sense, "incidentally" instead functions as an adverb. Correctly classifying cue phrases as discourse or sentential is critical for tasks that exploit discourse structure, e.g., anaphora resolution (Grosz & Sidner 1986).

While the problem of cue phrase classification has often been noted (Grosz & Sidner 1986; Halliday & Hassan 1976; Reichman 1985; Schiffrin 1987; Zuckerman & Pearl 1986), it has generally not received careful study. Recently, however, Hirschberg and Litman (1993) have presented rules for classifying cue phrases in both text and speech. Hirschberg and Litman pre-classified a set of naturally occurring cue phrases, described each cue phrase in terms of prosodic and textual features, then *manually* examined the data to construct rules that best predicted the classifications from the features.

This paper examines the utility of *machine learning* for automating the construction of rules for classifying cue phrases. A set of experiments are conducted that use two machine learning programs, CGRENDEL (Cohen 1992; 1993) and C4.5 (Quinlan 1986; 1987), to induce classification rules from sets of pre-classified cue phrases and their features. To support a quantitative and comparative evaluation of the automated and manual approaches, both the error rates and the content of the manually derived and learned rulesets are compared. The experimental results show that machine learning is indeed an effective technique for *automating* the generation of classification rules. The accuracy of the learned rulesets is often *higher than* the accuracy of the rules in (Hirschberg & Litman 1993), while the linguistic implications are more precise.

## Cue Phrase Classification

This section summarizes Hirschberg and Litman's study of the classification of multiple cue phrases in text and speech (Hirschberg & Litman 1993). The data from this study is used to create the input for the machine learning experiments, while the results are used as a benchmark for evaluating performance. The corpus examined was a technical address by a single speaker, lasting 75 minutes and consisting of approximately 12,500 words. The corpus yielded 953 instances of 34 different single word cue phrases. Hirschberg and Litman each classified the 953 tokens (as *discourse*, *sentential* or *ambiguous*) while listening to a recording and reading a transcription. Each token was also described as a set of *prosodic* and *textual* features. Previous observations in the literature correlating discourse structure with prosodic information, and discourse usages of cue phrases with initial position in a clause, contributed to the choice of features.

The prosody of the corpus was described using Pierrehumbert's theory of English intonation (Pierrehumbert 1980). In Pierrehumbert's theory, intonational contours are described as sequences of low (L) and high (H) *tones* in the *fundamental frequency (F0) contour* (the physical correlate of pitch). Intonational contours have as their domain the intonational phrase. A finite-state grammar describes the set of tonal sequences for an intonational phrase. A well-formed *intonational phrase* consists of one or more intermediate phrases followed by a boundary tone. A well-formed *intermediate phrase* has one or more pitch accents followed by a phrase accent. *Boundary tones* and *phrase accents* each consist of a single tone, while *pitch accents* con-sist of either a single tone or a pair of tones. There are two simple pitch accents (H* and L*) and four complex accents (L*+H, L+H*, H*+L, and H+L*). The * indicates which tone is aligned with the stressed syllable of the associated lexical item. Note that not every stressed syllable is accented. Lexical items that bear pitch accents are called *accented*, while those that do not are called *deaccented*.

Prosody was manually determined by examining the fundamental frequency (F0) contour, and by listening to the recording. To produce the F0 contour, the recording of the corpus was digitized and pitch-tracked using speech analysis software. This resulted in a display of the F0 where the x-axis represented time and the y-axis represented frequency in Hz. Various phrase final characteristics (e.g., phrase accents, boundary tones, as well as pauses and syllable lengthening) helped to identify intermediate and intonational phrases, while peaks or valleys in the display of the F0 contour helped to identify pitch accents.

In (Hirschberg & Litman 1993), every cue phrase was described using the following prosodic features. *Accent* corresponded to the pitch accent (if any) that was associated with the token. For both the intonational and intermediate phrases containing each token, the feature *composition of phrase* represented whether or not the token was *alone* in the phrase (the phrase contained only the token, or only cue phrases). *Position in phrase* represented whether the token was *first* (the first lexical item in the phrase – possibly preceded by other cue phrases), the last item in the phrase, or other.

Every cue phrase was also described in terms of the following textual features, derived directly from the transcript using fully automated methods. The *part of speech* of each token was obtained by running a program for tagging words with one of approximately 80 parts of speech on the transcript (Church 1988). Several characteristics of the token's immediate context were also noted, in particular, whether the token was immediately preceded or succeeded by *orthography* (punctuation or a paragraph boundary), and whether the token was immediately preceded or succeeded by a lexical item corresponding to a cue phrase.

The set of classified and described tokens was used to evaluate the accuracy of the classification models shown in Figure 1, developed in earlier studies. The prosodic model resulted from a study of 48 "now"s produced by multiple speakers in a radio call-in show (Hirschberg & Litman 1987). In a procedure similar to that described above, Hirschberg and Litman first classified and described each of the 48 tokens. They then examined their data manually to develop the prosodic model, which correctly classified all of the 48 tokens. (When later tested on 52 new examples of "now" from the radio corpus, the model also performed nearly perfectly). The model uniquely classifies any cue phrase using the features composition of

Prosodic Model:

**if** composition of intermediate phrase = alone **then** *discourse* (1)
**elseif** composition of intermediate phrase ≠ alone **then** (2)
    **if** position in intermediate phrase = first **then** (3)
        **if** accent = deaccented **then** *discourse* (4)
        **elseif** accent = L* **then** *discourse* (5)
        **elseif** accent = H* **then** *sentential* (6)
        **elseif** accent = complex **then** *sentential* (7)
    **elseif** position in intermediate phrase ≠ first **then**
        *sentential* (8)

Textual Model:

**if** preceding orthography = true **then** *discourse* (9)
**elseif** preceding orthography = false **then** *sentential* (10)

Figure 1: Decision tree representation of the classification models of (Hirschberg and Litman 1993).

intermediate phrase, position in intermediate phrase, and accent. When a cue phrase is uttered as a single intermediate phrase – possibly with other cue phrases (i.e., line (1) in Figure 1), or in a larger intermediate phrase with an initial position (possibly preceded by other cue phrases) and a L* accent or deaccented, it is classified as discourse. When part of a larger intermediate phrase and either in initial position with a H* or complex accent, or in a non-initial position, it is sentential. The textual model was also manually developed, and was based on an examination of the first 17 minutes of the single speaker technical address (Litman & Hirschberg 1990); the model correctly classified 89.4% of these 133 tokens. When a cue phrase is preceded by any type of orthography it is classified as discourse, otherwise as sentential.

The models were evaluated by quantifying their performance in correctly classifying two subsets of the 953 tokens from the corpus. The first subset (878 examples) consisted of only the *classifiable* tokens, i.e., the tokens that both Hirschberg and Litman classified as *discourse* or that both classified as *sentential*. The second subset, the *classifiable non-conjuncts* (495 examples), was created from the classifiable tokens by removing all examples of "and", "or" and "but". This subset was considered particularly reliable since 97.2% of non-conjuncts were classifiable compared to 92.1% of all tokens. The error rate of the prosodic model was 24.6% for the classifiable tokens and 14.7% for the classifiable non-conjuncts. The error rate of the textual model was 19.1% for the classifiable tokens and 16.1% for the classifiable non-conjuncts. In contrast, a model which just predicts the most frequent class in the corpus (sentential) has an error rate of 39% and 41% for the classifiable tokens and the classifiable non-conjuncts, respectively.

## Experiments using Machine Induction

This section describes experiments that use the machine learning programs C4.5 (Quinlan 1986; 1987) and CGRENDEL (Cohen 1992; 1993) to *automatically* in-duce cue phrase classification rules from both the data of (Hirschberg & Litman 1993) and an extension of this data. CGRENDEL and C4.5 are similar to each other and to other learning methods (e.g., neural networks) in that they induce rules from preclassified examples. Each program takes two inputs: 1) definitions of the classes to be learned, and of the names and values of a fixed set of features, and 2) the training data, i.e., a set of examples for which the class and feature values are specified. The output of each program is a set of classification rules, expressed in C4.5 as a decision tree and in CGRENDEL as an ordered set of if-then rules. Both CGRENDEL and C4.5 learn the classification rules using greedy search guided by an "information gain" metric.

The first set of experiments does not distinguish among the 34 cue phrases. In each experiment, a different subset of the features coded in (Hirschberg & Litman 1993) is examined. The experiments consider every feature in isolation (to comparatively evaluate the utility of each individual knowledge source for classification), as well as linguistically motivated sets of features (to gain insight into the interactions between the knowledge sources). The second set of experiments treats cue phrases individually. This is done by adding a lexical feature representing the cue phrase to each feature set from the first set of experiments. The potential use of such a lexical feature was noted but not used in (Hirschberg & Litman 1993). These experiments evaluate the utility of developing classification models specialized for particular cue phrases, and also provide qualitatively new linguistic insights into the data.

The first input to each learning program defines the classes and features. The classifications produced by Hirschberg and by Litman (*discourse*, *sentential*, and *ambiguous*) are combined into a single classification for each cue phrase. A cue phrase is classified as *discourse* (or as *sentential*) if both Hirschberg and Litman agreed upon the classification *discourse* (or upon *sentential*). A cue phrase is *non-classifiable* if at least one of Hirschberg and/or Litman classified the token as *ambiguous*, or one classified it as *discourse* while the other classified it as *sentential*. The features considered in the learning experiments are shown in Figure 2. Feature values can either be a numeric value or one of a fixed set of user-defined symbolic values. The feature representation shown here follows the representation of (Hirschberg & Litman 1993) except as noted. *Length of phrase* (P-L and I-L) represents the number of words in the phrase. This feature was not coded in the data from which the prosodic model was developed, but was coded (although not used) in the later data of (Hirschberg & Litman 1993). *Position in phrase* (P-P and I-P) uses numeric rather than symbolic values. The conjunction of the first two values for I-C is equivalent to *alone* in Figure 1. *Ambiguous*, the last value of A, is assigned when the prosodic anal-

- **Prosodic Features**
  - length of intonational phrase (P-L): integer.
  - position in intonational phrase (P-P): integer.
  - length of intermediate phrase (I-L): integer.
  - position in intermediate phrase (I-P): integer.
  - composition of intermediate phrase (I-C): only, only cue phrases, other.
  - accent (A): H*, L*, L*+H, L+H*, H*+L, H+L*, deaccented, ambiguous.
  - accent* (A*): H*, L*, complex, deaccented, ambiguous.
- **Textual Features**
  - preceding cue phrase (C-P): true, false, NA.
  - succeeding cue phrase (C-S): true, false, NA.
  - preceding orthography (O-P): comma, dash, period, paragraph, false, NA.
  - preceding orthography* (O-P*): true, false, NA.
  - succeeding orthography (O-S): comma, dash, period, false, NA.
  - succeeding orthography* (O-S*): true, false, NA.
  - part-of-speech (POS): article, coordinating conjunction, cardinal numeral, subordinating conjunction, preposition, adjective, singular or mass noun, singular proper noun, intensifier, adverb, verb base form, NA.
- **Lexical Feature**
  - token (T): actually, also, although, and, basically, because, but, essentially, except, finally, first, further, generally, however, indeed, like, look, next, no, now, ok, or, otherwise, right, say, second, see, similarly, since, so, then, therefore, well, yes.

Figure 2: Representation of features and their values, for use by C4.5 and CGRENDEL.

ysis of (Hirschberg & Litman 1993) is a disjunction (e.g., "H*+L or H*"). *NA* (not applicable) in the textual features reflects the fact that 39 recorded examples were not included in the transcription, which was done independently of (Hirschberg & Litman 1993). While the original representation noted the actual token (e.g., "and") when there was a preceding or succeeding cue phrase, here the value *true* encodes all such cases. Similarly, A*, O-P*, and O-S* re-represent the symbolic values of three features using a more abstract level of description (e.g., L*+H, L+H*, H*+L, and H+L* are represented as separate values in A but as a single value – the superclass *complex* – in A*). Finally, the lexical feature *token* is new to this study, and represents the actual cue phrase being described.

The second input to each learning program is training data, i.e., a set of examples for which the class and feature values are specified. Consider the following utterance, taken from the corpus of (Hirschberg & Litman 1993):

**Example 1** [(*Now*) (*now* that we have all been welcomed here)] it's time to get on with the business of the conference.

This utterance contains two cue phrases, corresponding to the two instances of "now". The brackets and parentheses illustrate the intonational and intermediate phrases, respectively, that contain the tokens. Note that a single intonational phrase contains both tokens, but that each token is uttered in a different interme-

Table 1: Multiple feature sets and their components.

|  | P-L | P-P | I-L | I-P | I-C | A | A* | C-P | C-S | O-P | O-P* | O-S | O-S* | POS |
|---|---|---|---|---|---|---|---|---|---|---|---|---|---|---|
| prosody | X | X | X | X | X | X | X | | | | | | | |
| hl93features | | | | X | X | X | X | | | | | | | |
| phrasing | X | X | X | X | X | | | | | | | | | |
| length | X | | X | | | | | | | | | | | |
| position | | X | | X | | | | | | | | | | |
| intonational | X | X | | | | | | | | | | | | |
| intermediate | | | X | X | X | | | | | | | | | |
| text | | | | | | | | X | X | X | X | X | X | X |
| adjacency | | | | | | | | X | X | | | | | |
| orthography | | | | | | | | | | X | X | X | X | |
| preceding | | | | | | | | X | | X | X | | | |
| succeeding | | | | | | | | | X | | | X | X | |
| speech-text | X | X | X | X | X | X | X | X | X | X | X | X | X | X |
| speech-adj | X | X | X | X | X | X | X | X | X | | | | | |

| P-L | P-P | I-L | I-P | I-C | A | A* | C-P | C-S | O-P | O-P* | O-S | O-S* | POS | T | Class |
|---|---|---|---|---|---|---|---|---|---|---|---|---|---|---|---|
| 9 | 1 | 1 | 1 | only | H*+L | complex | f | t | par. | t | f | f | adverb | now | discourse |
| 9 | 2 | 8 | 1 | other | H* | H* | t | f | f | f | f | f | adverb | now | sentential |

Figure 3: Representation of examples as features and their values.

diate phrase. If we were only interested in the feature P-L, the two examples would be represented in the training data as follows:

| P-L | Class |
|---|---|
| 9 | discourse |
| 9 | sentential |

The first column indicates the value assigned to the feature P-L, while the second column indicates how the example was classified. Thus, the length of the intonational phrase containing the first instance of "now" is 9 words, and the token is classified as a discourse usage.

In the first set of learning experiments, examples are represented using 28 different feature sets. First, there are 14 *single feature sets*, corresponding to each prosodic and textual feature. The example shown above illustrates how data is represented using the single feature set P-L. Second, there are 14 *multiple feature sets*, as described in Table 1. Each of these sets contains a linguistically motivated subset of at least 2 of the 14 prosodic and textual features. The first 7 sets use only prosodic features. *Prosody* considers all the prosodic features that were coded for each token. *Hl93features* considers only the coded features that were also used in the model shown in Figure 1. *Phrasing* considers all features of both the intonational and intermediate phrases containing the token (i.e., length of phrase, position of token in phrase, and composition of phrase). *Length* and *position* each consider only one of these features, but with respect to both the intonational and intermediate phrase. Conversely, *intonational* and *intermediate* each consider only one type of phrase, but consider all of the features. The next 5 sets use only textual features. *Text* considers all the textual features. *Adjacency* and *orthography* each consider a single textual feature, but consider both the preceding and succeeding immediate context. Conversely, *preceding* and *succeeding* consider contextual features relating to both orthography and cue phrases, but limit the context. The last two sets use both prosodic and textual features. *Speech-text* considers all features, while *speech-adj* does not consider orthography (which is subject to transcriber idiosyncrasy) and part of speech (which is dependent on orthography).

The second set of experiments considers 28 *tokenized feature sets*, constructed by adding *token* (the cue phrase being described) to each of the 14 single and 14 multiple feature sets. These sets will be referred to using the names of the single and multiple feature sets, concatenated with "+". Figure 3 illustrates how the two tokens in Example 1 would be represented using speech-text+. Consider the feature values for the first token. Since this token is the first lexical item in both the intonational and intermediate phrases which contain it, its position in both phrases (P-P and I-P) is 1. Since the intermediate phrase containing the token contains no other lexical items, its length (I-L) is 1 word and its composition (I-C) is *only* the token. The values for A and A* indicate that when the intonational phrase is described as a sequence of tones, the complex pitch accent H*+L is associated with the token. Finally, the utterance was transcribed such that it began a new paragraph. Thus the token was not preceded by another cue phrase (C-P), but it was preceded by a form of orthography (O-P and O-P*). Since the token was immediately followed by another instance of "now" in the transcription, the token was succeeded by another cue phrase (C-S) but was not succeeded by orthography (O-S and O-P*).

For each of the 56 feature sets (14 single feature,

| Set | Cgrendel | C4.5 | Set | Cgrendel | C4.5 | Set | Cgrendel | C4.5 | Set | Cgrendel | C4.5 |
|---|---|---|---|---|---|---|---|---|---|---|---|
| P-L | 32 | 32 | P-L+ | *21* | 31 | prosody | *15* | 16 | prosody+ | *16* | *15* |
| P-P | *16* | *16* | P-P+ | *16* | 18 | hl93features | 29 | 30 | hl93features+ | *23* | *28* |
| I-L | 25 | 25 | I-L+ | *20* | 26 | phrasing | *16* | *15* | phrasing+ | *14* | *15* |
| I-P | 25 | 25 | I-P+ | 25 | 26 | length | 26 | *24* | length+ | *18* | *24* |
| I-C | 36 | 36 | I-C+ | 27 | 36 | position | *18* | *18* | position+ | *15* | *17* |
| A | 28 | 40 | A+ | *19* | 40 | intonational | *17* | *16* | intonational+ | *15* | *16* |
| A* | 28 | 28 | A*+ | *18* | 26 | intermediate | *21* | *21* | intermediate+ | *18* | *22* |
| C-P | 40 | 40 | C-P+ | 28 | 39 | text | *18* | *18* | text+ | *18* | 20 |
| C-S | 41 | 40 | C-S+ | 28 | 39 | adjacency | 39 | 40 | adjacency+ | 28 | 39 |
| O-P | 20 | 40 | O-P+ | *17* | 35 | orthography | *18* | *18* | orthography+ | *17* | *19* |
| O-P* | *18* | *18* | O-P*+ | *17* | 20 | preceding | *18* | *18* | preceding+ | *17* | *19* |
| O-S | 34 | 35 | O-S+ | 26 | 31 | succeeding | 33 | 34 | succeeding+ | 25 | 32 |
| O-S* | 35 | 34 | O-S*+ | 27 | 32 | speech-text | *15* | *15* | speech-text+ | *16* | *13* |
| POS | 37 | 40 | POS+ | 27 | 34 | speech-adj | 30 | 29 | speech-adj+ | 27 | 28 |

Table 2: CGRENDEL and C4.5 error rates for the classifiable tokens (N=878).

14 multiple feature, and 28 token sets), 2 actual sets of examples are created as input to the learning systems. These sets correspond to the two subsets of the corpus examined in (Hirschberg & Litman 1993) – the classifiable tokens, and the classifiable non-conjuncts.

## Results

This section examines the results of running the learning programs C4.5 and CGRENDEL on 112 sets of examples (56 feature sets x 2 sets of examples). The results are qualitatively examined by comparing the linguistic content of the learned rulesets with the rules of Figure 1. The results are quantitatively evaluated by comparing the error rate of the learned rulesets in classifying new examples to the error rate of the rules of Figure 1. The *error rate* of a set of rules is computed by using the rules to predict the classification of a set of (pre-classified) examples, then comparing the predicted and known classifications. In the cue phrase domain, the error rate is computed by summing the number of discourse examples misclassified as sentential with the number of sentential examples misclassified as discourse, then dividing by the total number of examples. *Cross-validation* (Weiss & Kulikowski 1991) is used to estimate the error rates of the learned rulesets. Instead of running each learning program once on each of the 112 sets of examples, 10 runs are performed, each using a random 90% of the examples for *training* (i.e., for learning the ruleset) and the remaining 10% for *testing*. An estimated error rate is obtained by averaging the error rate on the testing portion of the data from each of the 10 runs. Note that for each run, the training and testing examples are disjoint subsets of the same set of examples, and the training set is much larger than the test set. In contrast (as discussed above), the "training" and test sets for the intonational model of (Hirschberg & Litman 1993) were taken from different corpora, while for the textual model of (Hirschberg & Litman 1993) the test set was a superset of the training set. Furthermore, more data was used for testing than for training, and the computation of the error rate did not use cross-validation.

Table 2 presents the estimated error of the learned rulesets on the 878 classifiable examples in the corpus. Each numeric cell shows the result (as a percentage) for one of the 56 feature sets. The standard error for each cell ranged from .6 to 2.7. The left half of the table considers the single feature and single feature plus token sets, while the right half considers the multiple features with and without token. The top of the table considers prosodic features, the bottom textual features, and the bottom right prosodic/textual combinations. The error rates in italics indicate that the performance of the learned ruleset exceeds the performance reported in (Hirschberg & Litman 1993), where the rules of Figure 1 were tested using 100% of the 878 classifiable tokens. These error rates were 24.6% and 19.1% for the intonational and textual models, respectively.

When considering only a single intonational feature (the first 3 columns of the first 7 rows), the results of the learning programs suggest that position in intonational phrase (P-P) is the most useful feature for cue phrase classification. In addition, this feature classifies cue phrases significantly better than the 3 feature prosodic model of Figure 1. The majority of the learned multiple feature rulesets (columns 7-9) also perform better than the model of Figure 1, although none significantly improve upon the single feature ruleset. Note that the performance of the manually derived model is better than the performance of *hl93features* (which uses the same set of features but in different rules). In fact, *hl93features* has among the worst performance of any of the learned prosodic rulesets. This suggests that the prosodic feature set most useful for classifying "now" did not generalize to other cue phrases. The ease of exploring large training sets and regenerating rules for new training data appear to be significant advantages of the automated approach.

An examination of the learned rulesets shows that they are quite comparable in content to relevant por-

Ruleset learned from P-P using C4.5:

**if** position in intonational phrase $\leq 1$ **then** *discourse*
**elseif** position in intonational phrase $> 1$ **then** *sentential*

Ruleset learned from P-P using CGRENDEL:

**if** position in intonational phrase $\geq 2$ **then** *sentential*
default is on *discourse*

Ruleset learned from prosody using C4.5:

**if** position in intonational phrase $\leq 1$ **then** *discourse*
**elseif** position in intonational phrase $> 1$ **then**
   **if** length of intermediate phrase $\leq 1$ **then** *discourse*
   **elseif** length of intermediate phrase $> 1$ **then** *sentential*

Ruleset learned from prosody using CGRENDEL:

**if** (position in intonational phrase $\geq 2$) $\wedge$
  (length of intermediate phrase $\geq 2$) **then** *sentential*
**if** ($7 \geq$ position in intonational phrase $\geq 4$) $\wedge$
  (length of intonational phrase $\geq 10$) **then** *sentential*
**if** (length of intermediate phrase $\geq 2$) $\wedge$
  (length of intonational phrase $\leq 7$) $\wedge$
  (accent = H*) **then** *sentential*
**if** (length of intermediate phrase $\geq 2$) $\wedge$
  (length of intonational phrase $\leq 9$) $\wedge$
  (accent = H*+L) **then** *sentential*
**if** (length of intermediate phrase $\geq 2$) $\wedge$
  (accent = deaccent) **then** *sentential*
**if** (length of intermediate phrase $\geq 8$) $\wedge$
  (length of intonational phrase $\leq 9$) $\wedge$
  (accent = L*) **then** *sentential*
default is on *discourse*

Ruleset learned from O-P* using C4.5:

**if** preceding orthography* = NA **then** *discourse*
**elseif** preceding orthography* = false **then** *sentential*
**elseif** preceding orthography* = true **then** *discourse*

Ruleset learned from O-P* using CGRENDEL:

**if** preceding orthography* = false **then** *sentential*
default is on *discourse*

Ruleset learned from speech-text+ using C4.5:

**if** position in intonational phrase $<= 1$ **then**
  **if** preceding orthography* = NA **then** *discourse*
  **elseif** preceding orthography* = true **then** *discourse*
  **elseif** preceding orthography* = false **then**
    **if** length of intermediate phrase $> 12$ **then** *discourse*
    **elseif** length of intermediate phrase $\leq 12$ **then**
      **if** length of intermediate phrase $\leq 1$ **then** *discourse*
      **elseif** length of intermediate phrase $> 1$ **then** *sentential*
**elseif** position in intonational phrase $> 1$ **then**
  **if** length of intermediate phrase $\leq 1$ **then** *discourse*
  **elseif** length of intermediate phrase $> 1$ **then** *sentential*

Figure 4: Example rulesets learned from different feature sets (classifiable tokens).

tions of Figure 1, and often contain further linguistic insights. Consider the rulesets learned from P-P, shown in the top of Figure 4. C4.5 represents its learned ruleset using a decision tree, while CGRENDEL instead produces a set of if-then rules. When multiple rules are applicable, CGRENDEL applies a conflict resolution strategy; when no rules are applicable, the default (the last statement) is used. Both programs produce both unsimplified and pruned rulesets. Only the simplified rulesets are considered in this paper. Both of the learned rulesets say that if the token is not in the initial position of the intonational phrase, classify as *sentential*; otherwise classify as *discourse*. Note the correspondence with line (8) in Figure 1. Figure 4 also illustrates the more complex rulesets learned using the larger set of features in *prosody*. The C4.5 model is similar to lines (1), (3) and (8) of Figure 1. (Note that the length value 1 is equivalent to the composition value *only*.) In the CGRENDEL hypothesis, the first 2 rules correlate sentential status with (among other things) non-initial position, and the second 2 with H* and H*+L accents; these rules are similar to rules (6)-(8) in Figure 1. However, the last 2 CGRENDEL rules also correlate L* and no accent with sentential status when the phrase is of a certain length, while rules (4) and (5) in Figure 1 provide a different interpretation and do not take length into account. (Recall that length was coded by Hirschberg and Litman only in their test data. Length was thus never used to generate or revise their prosodic model.) Both of the learned rulesets perform similarly to each other, and outperform the prosodic model of Figure 1.

Examination of the learned textual rulesets yields similar findings. Consider the rulesets learned from O-P* (preceding orthography, where the particular type of orthography is not noted), shown towards the bottom of Figure 4. These rules outperform the rules using the other single features, and perform comparably to the model in Figure 1 and to the multiple feature textual rulesets incorporating preceding orthography. Again, note the similarity to lines (9) and (10) of Figure 1. Also note that the textual feature values were obtained directly from the transcript, while determining the values of prosodic features required manual analysis.

Performance of a feature set is often improved when the additional feature *token* is taken into account (columns 4-6 and 10-12). This phenomenon will be discussed below. Finally, *speech-text* and *speech-text+*, which consider every available feature, outperform the manually and nearly all the automatically derived models. The last example in Figure 4 is the best performing ruleset in Table 2, the C4.5 hypothesis learned from *speech-text+*.

Table 3 presents the results using a smaller portion of the corpus, the 495 classifiable non-conjuncts. The error rates of the intonational and textual models of Figure 1 on this subcorpus decrease to 14.7% and 16.1%, respectively (Hirschberg & Litman 1993). Without the feature *token*, the single feature sets based on position and preceding orthography are again the best performers, and along with many multiple feature non-token sets, perform nearly as well as the models in Figure 1.

When the feature *token* is taken into account, however, the learned rulesets outperform the models of (Hirschberg & Litman 1993) (which did not con-

| Set | Cgrendel | C4.5 | Set | Cgrendel | C4.5 | Set | Cgrendel | C4.5 | Set | Cgrendel | C4.5 |
|---|---|---|---|---|---|---|---|---|---|---|---|
| P-L | 33 | 32 | P-L+ | 17 | 31 | prosody | 17 | 19 | prosody+ | 15 | 16 |
| P-P | 18 | 18 | P-P+ | 14 | 19 | hl93features | 18 | 18 | hl93features+ | 17 | 18 |
| I-L | 25 | 25 | I-L+ | 16 | 25 | phrasing | 19 | 18 | phrasing+ | 12 | 17 |
| I-P | 19 | 19 | I-P+ | 17 | 18 | length | 27 | 26 | length+ | 16 | 24 |
| I-C | 35 | 35 | I-C+ | 18 | 32 | position | 19 | 19 | position+ | 13 | 17 |
| A | 30 | 29 | A+ | 12 | 29 | intonational | 20 | 18 | intonational+ | 16 | 19 |
| A* | 28 | 28 | A*+ | 15 | 31 | intermediate | 19 | 21 | intermediate+ | 16 | 18 |
| C-P | 40 | 39 | C-P+ | 16 | 33 | text | 19 | 20 | text+ | 12 | 15 |
| C-S | 39 | 39 | C-S+ | 17 | 39 | adjacency | 40 | 40 | adjacency+ | 15 | 43 |
| O-P | 17 | 18 | O-P+ | 10 | 14 | orthography | 18 | 17 | orthography+ | 13 | 18 |
| O-P* | 17 | 17 | O-P*+ | 12 | 15 | preceding | 17 | 19 | preceding+ | 13 | 16 |
| O-S | 30 | 31 | O-S+ | 18 | 31 | succeeding | 30 | 30 | succeeding+ | 18 | 31 |
| O-S* | 32 | 31 | O-S*+ | 16 | 32 | speech-text | 14 | 16 | speech-text+ | 16 | 17 |
| POS | 38 | 41 | POS+ | 17 | 31 | speech-adj | 17 | 18 | speech-adj+ | 18 | 21 |

Table 3: CGRENDEL and C4.5 error rates for the classifiable non-conjuncts (N=495).

Ruleset learned from A+ using CGRENDEL:

**if** accent = L* **then** *discourse*
**if** (accent = deaccent) ∧ (token = say) **then** *discourse*
**if** (accent = deaccent) ∧ (token = so) **then** *discourse*
**if** (accent = L+H*) ∧ (token = further) **then** *discourse*
**if** (accent = L+H*) ∧ (token = indeed) **then** *discourse*
**if** token = now **then** *discourse*
**if** token = finally **then** *discourse*
**if** token = however **then** *discourse*
**if** token = ok **then** *discourse*
default is on *sentential*

Figure 5: Using the feature *token* during learning (classifiable non-conjuncts).

sider this feature), and also provide new insights into cue phrase classification. Figure 5 shows the CGRENDEL ruleset learned from A+, which reduces the 30% error rate of A to 12%. The first rule corresponds to line (5) of Figure 1. In contrast to line (4), however, CGRENDEL uses deaccenting to predict *discourse* for only the tokens "say" and "so." If the token is "now", "finally", "however", or "ok", *discourse* is assigned (for all accents). In all other deaccented cases, *sentential* is assigned (using the default). Similarly, in contrast to line (7), the complex accent L+H* predicts *discourse* for the cue phrases "further" or "indeed" (and also for "now", "finally", "however" and "ok"), and *sentential* otherwise. Rulesets such as these suggest that even though features such as accent may not characterize all cue phrases, they may nonetheless be used successfully if the feature is used differently for different cue phrases or subsets of cue phrases.

Note that in the subcorpus of non-conjuncts (in contrast to the classifiable subcorpus), machine learning only improves on human performance by considering more features, either the extra feature *token* or textual and prosodic features in combination. This might reflect the fact that the manually derived theories already achieve optimal performance with respect to the examined features in this less noisy subcorpus, and/or that the automatically derived theory for this subcorpus was based on a smaller training set than used in the previous subcorpus.

## Related Work in Discourse Analysis

Grosz and Hirschberg (1992) used the system CART (Brieman *et al.* 1984) to construct decision trees for classifying aspects of discourse structure from intonational feature values. Siegel (in press) was the first to apply machine learning to cue phrases. He developed a genetic learning algorithm to induce decision trees using the non-ambiguous examples of (Hirschberg & Litman 1993) (using the classifications of only one judge) as well as additional examples. Each example was described using a feature corresponding to *token*, as well as textual features containing the lexical or orthographic item immediately to the left of and in the 4 positions to the right of the example. Thus, new textual features were examined. Prosodic features were not investigated. Siegel reported a 21% estimated error rate, with half of the corpus used for training and half for testing. An examination of Table 2 shows that the error of the best C4.5 and CGRENDEL rulesets was often lower than 21% (even for theories which did not consider the token), as was the 19.1% error of the textual model of (Hirschberg & Litman 1993). Siegel and McKeown (1994) have also proposed a method for developing linguistically viable rulesets, based on the partitioning of the training data produced during induction.

## Conclusion

This paper has demonstrated the utility of *machine learning* techniques for cue phrase classification. A first set of experiments were presented that used the programs CGRENDEL (Cohen 1992; 1993) and C4.5 (Quinlan 1986; 1987) to induce classification rules from the preclassified cue phrases and their features that were used as test data in (Hirschberg & Litman 1993). The results of these experiments suggest that machine learning is an effective technique for not only *automating* the generation of linguistically plausible classifica-tion rules, but also for *improving* accuracy. In particular, a large number of learned rulesets (including P-P, an extremely simple one feature model) had significantly lower error rates than the rulesets of (Hirschberg & Litman 1993). One possible explanation is that the hand-built classification models were derived using very small "training" sets; as new data became available, this data was used for testing but not for updating the original models. In contrast, machine learning supported the building of rulesets using a much larger amount of the data for training. Furthermore, if new data becomes available, it is trivial to regenerate the rulesets. For example, in a second set of experiments, new classification rules were induced using the feature *token*, which was not considered in (Hirschberg & Litman 1993). Allowing the learning programs to treat cue phrases individually further improved the accuracy of the resulting rulesets, and added to the body of linguistic knowledge regarding cue phrases.

Another advantage of the machine learning approach is that the ease of inducing rulesets from many different sets of features supports an exploration of the comparative utility of different knowledge sources. For example, when prosodic features were considered in isolation, only position in intonational phrase appeared to be useful for classification. However, in combination with the token, several additional prosodic features appeared to be equally useful. The results of this paper suggest that machine learning is a useful tool for cue phrase classification, when the amount of data is too large for human analysis, and/or when an analysis goal is to gain a better understanding of the different aspects of the data.

## Acknowledgments

I would like to thank William Cohen and Jason Catlett for help in using CGRENDEL and C4.5, and William Cohen, Ido Dagan, Julia Hirschberg, and Eric Siegel for comments on an earlier version of this paper.